\documentclass[pdflatex,sn-mathphys-num]{sn-jnl}

\usepackage{graphicx}%
\usepackage{multirow}%
\usepackage{amsmath,amssymb,amsfonts}%
\usepackage{amsthm}%
\usepackage{mathrsfs}%
\usepackage[title]{appendix}%
\usepackage{xcolor}%
\usepackage{textcomp}%
\usepackage{manyfoot}%
\usepackage{booktabs}%
\usepackage{algorithm}%
\usepackage{algorithmicx}%
\usepackage{algpseudocode}%
\usepackage{listings}%

\usepackage{subcaption}%
\usepackage{xurl}%
\usepackage{enumitem}%

\usepackage{amsopn}%
\newcommand{\matr}[1]{#1}

\makeatletter
\newcommand{\subalign}[1]{%
  \vcenter{%
    \Let@ \restore@math@cr \default@tag
    \baselineskip\fontdimen10 \scriptfont\tw@
    \advance\baselineskip\fontdimen12 \scriptfont\tw@
    \lineskip\thr@@\fontdimen8 \scriptfont\thr@@
    \lineskiplimit\lineskip
    \ialign{\hfil$\m@th\scriptstyle##$&$\m@th\scriptstyle{}##$\hfil\crcr
      #1\crcr
    }%
  }%
}
\makeatother

\newcommand{\tr}{^\top\! }

\newcommand{\argmin}[1]{\underset{#1}{\operatorname{argmin}}}

\newcommand{\Trace}{\mathrm{Tr}}

\newcommand{\veccc}[2][]{\vecc_\mathrm{#1}\left({#2}\right)}

\newcommand{\Gr}{\mathrm{Gr}}

\newcommand{\vecc}{\mathrm{vec}}

\newcommand{\calD}{\mathcal{D}}

\newcommand{\norm}[2][2]{\left\|{#2}\right\|_\mathrm{#1}}
\newcommand{\sqnorm}[2][2]{\norm[#1]{#2}^2}

\newcommand{\frobnorm}[2][F]{\left\|{#2}\right\|_\mathrm{#1}}
\newcommand{\sqfrobnorm}[2][F]{\frobnorm[#1]{#2}^2}

\newcommand{\TODOF}[1]{}

\usepackage{cleveref}%
\AtBeginDocument{}%

\theoremstyle{thmstyleone}%

\theoremstyle{thmstyletwo}%

\theoremstyle{thmstylethree}%

\begin{document}

\title[Graph-Regularized Low-Rank Matrix Completion by Variable Projection]{Graph-Regularized Low-Rank Matrix Completion by Variable Projection}

\author*[1]{\fnm{Beno\^{i}t} \sur{Loucheur}}\email{benoit.loucheur@uclouvain.be}

\author[1]{\fnm{P.-A.} \sur{Absil}}\email{pa.absil@uclouvain.be}

\author[2]{\fnm{Michel} \sur{Journ\'{e}e}}\email{michel.journee@meteo.be}

\affil*[1]{\orgdiv{ICTEAM Institute}, \orgname{UCLouvain}, \orgaddress{\postcode{1348} \city{Louvain-la-Neuve}, \country{Belgium}}}

\affil[2]{\orgdiv{Department of Climatology}, \orgname{Royal Meteorological Institute of Belgium}, \orgaddress{\postcode{1180} \city{Uccle}, \country{Belgium}}}


\abstract{We address the low-rank matrix completion problem by incorporating graph regularization into the existing Riemannian Trust-Region Matrix Completion (RTRMC) framework. The latter uses the geometry of the low-rank constraint to remodel the problem as an unconstrained optimization problem on a single Grassmann manifold. Our approach, named Graph-Regularized RTRMC (GR-RTRMC), exploits the inherent relationships between rows and columns of the matrix. By using these relationships, we aim to improve the accuracy and robustness of matrix completion, particularly in scenarios where the underlying data exhibits strong correlations between rows or columns.}



\maketitle


\section{Introduction}
In many application domains, such as recommendation systems~\cite{netflix}, weather forecasting~\cite{weather_forecasting_missing1}, and network analysis~\cite{network_missing1,network_missing2}, the available data is often incomplete due to missing measurements, transmission failures, or other technical limitations~\cite{dataloss}. These gaps in the data make it challenging to analyze and utilize the information effectively, requiring robust approaches to estimate the missing values.

Low-rank matrix completion has emerged as a widely used method to address this issue. It is based on the assumption that the observed data can be represented as a low-rank matrix, enabling the reconstruction of missing entries from a partially observed subset.

Throughout this paper, we denote by $\matr{M} \in \mathbb{R}^{m \times n}$ the complete target matrix that we aim to recover. In practice, we only observe a subset of its entries at indices $\Omega \subseteq \{1,\ldots,m\} \times \{1,\ldots,n\}$.

\subsection{Related work}
Low-rank matrix completion is classically formulated as the problem of finding a low-rank matrix that satisfies constraints on the observed entries:
\begin{equation}
	\min_{\matr{X}\in\mathbb{R}^{m\times n}} \quad  \mathrm{rank}(\matr{X}) \qquad \textrm{s.t.} \quad  \mathcal{P}_\Omega(\matr{X}) = \mathcal{P}_\Omega(\matr{M}).
	\label{eq:RM}
\end{equation}
Here, $\mathcal{P}_\Omega$ is the projection operator that restricts a matrix to its observed entries:
\begin{equation}
[\mathcal{P}_\Omega(\matr{X})]_{ij} = \begin{cases}
\matr{X}_{ij} & \text{if } (i,j) \in \Omega \\
0 & \text{otherwise}.
\end{cases}
\end{equation}
Despite its intuitive formulation, this problem is NP-hard~\cite{CandesRecht_article} due to the non-convexity of the rank function, making it computationally infeasible for large-scale matrices.

To circumvent the challenges posed by rank minimization, a convex relaxation replaces the rank function with the nuclear norm, defined as the sum of the singular values of the matrix. The relaxed problem is expressed as~\cite{CandesRecht_article}:
\begin{equation}
	\min_{\matr{X}\in\mathbb{R}^{m\times n}} \quad \|\matr{X}\|_\ast \qquad \textrm{s.t.} \quad \mathcal{P}_\Omega(\matr{X}) = \mathcal{P}_\Omega(\matr{M}),
\end{equation}
where \( \|\matr{X}\|_\ast \) serves as a convex surrogate for the rank. While effective for small matrices, nuclear norm minimization scales poorly with the matrix dimensions due to its reliance on repeated singular value decomposition (SVD) during optimization.

To improve computational efficiency, many methods adopt a factorized representation. These methods approximate the matrix as \( \matr{X} = \matr{U}\matr{W} \), where \( \matr{U} \in \mathbb{R}^{m \times r} \) and \( \matr{W} \in \mathbb{R}^{r \times n} \). This reformulation enforces a rank constraint implicitly, transforming the problem into~\cite{10.1145/1102351.1102441}:
\begin{equation}
	\min_{\substack{\matr{U}\in\mathbb{R}^{m\times r}\\\matr{W}\in\mathbb{R}^{r\times n}}} \quad \|\mathcal{P}_\Omega(\matr{U}\matr{W}) - \mathcal{P}_\Omega(\matr{M})\|_F^2 + \lambda(\|\matr{U}\|_F^2 + \|\matr{W}\|_F^2),
	\label{eq:MC2}
\end{equation}
where \( \lambda \) is a regularization parameter that controls the trade-off between fitting the observed entries and penalizing the complexity of \( \matr{U} \) and \( \matr{W} \). This factorized formulation significantly reduces the computational cost, making it suitable for large-scale problems~\cite{wang2025approximation}.

However, the standard factorized approach treats all rows and columns independently, ignoring potential structural relationships in the data. In many real-world applications, additional side information about the relationships between rows and columns is available. This information is often naturally encoded as graphs that describe relationships, such as user similarities in recommendation systems or spatial proximity in weather data. 

To leverage this structural information, graph-based regularization can be incorporated into matrix completion models to enforce smoothness along graph structures~\cite{graph_form, dong2021riemannian, mongia2019multiple, elmahdy2020hierarchical}. This means that connected nodes in the graph (e.g., nearby weather stations or similar users) should have similar latent representations. This leads to the following formulation~\cite{grals}:
\begin{equation}
\min_{\substack{\matr{U}\in\mathbb{R}^{m\times r}\\\matr{W}\in\mathbb{R}^{r\times n}}} 
\Big\{ \frac{1}{2}\sqfrobnorm[2]{\mathcal{P}_\Omega(\matr{U}\matr{W})-\mathcal{P}_\Omega(\matr{M})} + \frac{\lambda_\mathrm{w}}{2}\sqfrobnorm{U}
+ \frac{\lambda_\mathrm{h}}{2}\sqfrobnorm{W} + \frac{\lambda_\mathrm{L}}{2}\sqnorm[\calD,u]{U} + \frac{\lambda_\mathrm{L}}{2}\sqnorm[\calD,w]{W} \Big\}.
\end{equation}
Here, the Dirichlet semi-norm \( \|\cdot\|_\mathcal{D} \) encodes graph smoothness constraints. For the row graph with Laplacian matrix \( \matr{L}_\mathrm{u} \in \mathbb{R}^{m \times m} \), we have:
\begin{equation}
\|\matr{U}\|_{\mathcal{D},\mathrm{u}}^2 = \text{Tr}(\matr{U}^T \matr{L}_\mathrm{u} \matr{U}) = \sum_{k=1}^r \matr{u}_k^T \matr{L}_\mathrm{u} \matr{u}_k,
\end{equation}
where \( \matr{u}_k \) is the \( k \)-th column of \( \matr{U} \). Since \( \matr{L}_\mathrm{u} = \matr{D}_\mathrm{u} - \matr{A}_\mathrm{u} \) with adjacency matrix \( \matr{A}_\mathrm{u} \) and degree matrix \( \matr{D}_\mathrm{u} \), this expands to:
\begin{equation}
\|\matr{U}\|_{\mathcal{D},\mathrm{u}}^2 = \sum_{k=1}^r \sum_{(i,j) \in E_\mathrm{u}} A_{\mathrm{u},ij} (U_{ik} - U_{jk})^2.
\end{equation}
This formulation penalizes large differences between connected nodes, encouraging similar rows (those connected in the graph) to have similar latent representations. The column regularization \( \|\matr{W}\|_{\mathcal{D},\mathrm{w}}^2 \) follows analogously with the column graph Laplacian \( \matr{L}_\mathrm{w} \). Methods such as Graph-Regularized Alternating Least Squares (GRALS)~\cite{grals} solve this formulation using the alternating minimization scheme.

Another promising line of work explores manifold-based optimization. The Riemannian Trust-Region Matrix Completion (RTRMC)~\cite{RTRMC_ABSIL_BOUMAL, BouAbs2015} framework operates on the Grassmann manifold \( \Gr(m, r) \), i.e., the set of \( r \)-dimensional subspaces in \( \mathbb{R}^m \). To handle the observed entries constraint, RTRMC employs a weighted formulation using a confidence matrix \( \matr{C} \) and the Hadamard product \( \odot \). This approach differs from the projection operator \( \mathcal{P}_\Omega \) used in previous formulations. Specifically, let us define the \( \Omega \)-norm of a matrix \( \matr{M} \) as:
\begin{equation}
\|\matr{M}\|_\Omega^2 := \sum_{(i,j) \in \Omega} M_{ij}^2,
\end{equation}
where \( \Omega \) denotes the set of observed entries. The confidence matrix \( \matr{C} \) assigns weights to observed entries, with \( C_{ij} > 0 \) for \( (i,j) \in \Omega \) and \( C_{ij} = 0 \) otherwise. RTRMC solves the following optimization problem:
\begin{equation}
        \min_{\mathcal{U}\in\Gr(m,r)}\Big(\min_{\matr{W}\in\mathbb{R}^{r\times n}} \frac{1}{2}\|\matr{C}\odot (\matr{U}\matr{W}-\matr{M})\|_\Omega^2 + \frac{\lambda^2}{2} \|\mathcal{P}_{\bar{\Omega}}(\matr{U}\matr{W})\|_F^2\Big).
    \label{eqn:rtrmc}
\end{equation}
\subsection{Contribution}
Building on RTRMC, we propose a graph-regularized extension that combines the advantages of Riemannian optimization with graph-based regularization. Our key contribution is extending the RTRMC framework by incorporating Dirichlet semi-norms that enforce smoothness along graph-encoded relationships. The proposed method solves the following optimization problem:
\begin{equation}
\min_{\mathcal{U}\in\Gr(m,r)}\Big(\min_{\matr{W}\in\mathbb{R}^{r\times n}} 
        \frac{1}{2}\|\matr{C}\odot (\matr{U}\matr{W}-\matr{M})\|_\Omega^2
         + \frac{\lambda^2}{2} \|\mathcal{P}_{\bar{\Omega}}(\matr{U}\matr{W})\|_F^2
         + \frac{\lambda_\mathrm{u}}{2}\|\matr{U}\|_{\mathcal{D},\mathrm{u}}^2
        + \frac{\lambda_\mathrm{w}}{2}\|\matr{W}\|_{\mathcal{D},\mathrm{w}}^2\Big),
    \label{eq:proposed}
\end{equation}
where the graph regularization terms \(\|\matr{U}\|_{\mathcal{D},\mathrm{u}}^2\) and \(\|\matr{W}\|_{\mathcal{D},\mathrm{w}}^2\) penalize deviations along graph structures associated with rows and columns respectively.


To demonstrate the effectiveness of our proposed method, we conduct experiments on meteorological datasets from Belgium and France, as well as the MovieLens collaborative filtering dataset. Our evaluation framework uses k-fold cross-validation to ensure robustness and fair comparisons across methods. The results highlight the superior performance of the graph-regularized RTRMC model, showing improved accuracy in recovering missing entries compared to baseline methods. These findings highlight the value of using graph-based structure when filling in missing data in matrices, especially when there are relationships between rows and columns.

A brief preview of this work, centered on the Belgian weather data application, was presented at the BNAIC/BeNeLearn 2024 conference~\cite{loucheur2024}. The present paper extends the conference version with: (i) a complete theoretical and algorithmic development of the proposed model, (ii) experiments on larger-scale French meteorological data and on the MovieLens collaborative filtering dataset, and (iii) detailed case studies analyzing failure modes during extreme weather events.

The proposed method is presented in Section~\ref{sec:model}. Numerical experiments are reported in Section~\ref{sec:experiments}. Conclusions are drawn in Section~\ref{sec:conclusions}.

\section{Graph-Regularized Riemannian Trust-Region Matrix Completion}
\label{sec:model}
We present the Graph-Regularized Riemannian Trust-Region Matrix Completion \linebreak(GR-RTRMC) method, which enhances the classical RTRMC approach by incorporating graph-based regularization. This extension leverages structural relationships between rows and columns of the target matrix to improve recovery performance, particularly in applications where such relationships naturally arise.

Building upon the RTRMC framework introduced in \eqref{eqn:rtrmc}, we incorporate the graph regularization terms from \eqref{eq:proposed} to formulate our optimization problem over matrix factors $\matr{U} \in \mathbb{R}_*^{m \times r}$ and $\matr{W} \in \mathbb{R}^{r \times n}$, where $\mathbb{R}_*^{m \times r}$ denotes the set of full-rank $m \times r$ matrices. We use the confidence matrix $\matr{C}$ (with $\matr{C}_{ij} > 0$ for $(i,j) \in \Omega$ and $\matr{C}_{ij} = 0$ otherwise), graph Laplacians $\matr{L}_\mathrm{u} \in \mathbb{R}^{m \times m}$ and $\matr{L}_\mathrm{w} \in \mathbb{R}^{n \times n}$ for rows and columns respectively, and $\overline{\Omega} = \{1,\ldots,m\} \times \{1,\ldots,n\} \setminus \Omega$ to denote the unobserved entries:

\begin{align}
\min_{\matr{U},\matr{W}} h(\matr{U},\matr{W}) &= \frac{1}{2}\|\matr{C} \odot (\matr{U}\matr{W} - \matr{M})\|^2_\Omega + \frac{\lambda^2}{2}\|\matr{U}\matr{W}\|^2_{\overline{\Omega}} \nonumber \\
&\quad + \frac{\lambda_\mathrm{u}}{2}\text{Trace}(\matr{U}^T\matr{L}_\mathrm{u}\matr{U}) + \frac{\lambda_\mathrm{w}}{2}\text{Trace}(\matr{W}\matr{L}_\mathrm{w}\matr{W}^T).
\label{eq:gr-objective}
\end{align}

The term weighted by $\lambda > 0$ regularizes unobserved entries, while $\lambda_\mathrm{u}, \lambda_\mathrm{w} \geq 0$ control graph regularization strength. The graph regularization terms encourage similar rows and columns (as defined by the respective graph structures) to have similar representations in the factorization.

\subsection{Optimization on the Grassmann Manifold}
The key insight~\cite{RTRMC_ABSIL_BOUMAL, BouAbs2015} enabling efficient optimization of~\eqref{eq:gr-objective} lies in recognizing that the factorization $\matr{U}\matr{W}$ exhibits a fundamental invariance property. This observation allows us to reformulate the high-dimensional optimization problem over matrix factors into a more structured optimization problem on a Riemannian manifold.

To understand this reformulation, we begin by noting that for any fixed left factor $\matr{U} \in \mathbb{R}_{*}^{m \times r}$, finding the optimal right factor $\matr{W}$ reduces to solving a regularized least-squares problem:
\begin{equation}
    \matr{W}_{\matr{U}} = \argmin{\matr{W} \in \mathbb{R}^{r \times n}} h(\matr{U},\matr{W}).
    \label{eq:wu-definition}
\end{equation}

By substituting this optimal $\matr{W}_{\matr{U}}$ back into the original objective \eqref{eq:gr-objective}, we obtain the reduced cost function:
\begin{equation}
f(\matr{U}) = h(\matr{U}, \matr{W}_{\matr{U}}),
\label{eq:reduced-cost}
\end{equation}
a technique known as~\emph{variable projection}~\cite{10.1007/s10589-012-9492-9}. The crucial observation is that $W_U$ depends only on the column space of $U$, not on the specific choice of basis for that subspace. To see this, consider any invertible transformation $M \in \mathbb{R}_*^{r \times r}$ and define $\tilde{U} = UM$. Then the optimal factor for $\tilde{U}$ satisfies:
\begin{equation}
W_{\tilde{U}} = W_{UM} = M^{-1}W_U.
\end{equation}

This relationship ensures that the matrix product remains invariant under basis changes:
\begin{equation}
\tilde{U}W_{\tilde{U}} = (UM)(M^{-1}W_U) = UW_U.
\end{equation}

This invariance property is fundamental: it shows that the quality of any factorization $UW$ depends only on the column space spanned by $U$, not on the particular orthonormal basis chosen to represent it. This naturally motivates a formulation on the Grassmann manifold $\text{Gr}(m,r)$, which is precisely the set of $r$-dimensional subspaces of $\mathbb{R}^m$.

For computational efficiency, we work with orthonormal representations on the Stiefel manifold $\text{St}(m,r) = \{\matr{U} \in \mathbb{R}^{m \times r} : \matr{U}^T\matr{U} = I_r\}$. When $\matr{U}$ has orthonormal columns, we can exploit the identity $\|\matr{U}\matr{W}_{\matr{U}}\|^2_{\overline{\Omega}} = \|\matr{W}_{\matr{U}}\|^2_F - \|\matr{U}\matr{W}_{\matr{U}}\|^2_\Omega$ (valid due to $\|\matr{U}\matr{W}_{\matr{U}}\|^2_F = \|\matr{W}_{\matr{U}}\|^2_F$ when $\matr{U}^T\matr{U} = I_r$) to rewrite our objective as:

\begin{align}
f(\matr{U}) = \frac{1}{2}\|\matr{C} \odot (\matr{U}\matr{W}_{\matr{U}} - \matr{M})\|^2_\Omega &+ \frac{\lambda^2}{2}(\|\matr{W}_{\matr{U}}\|^2_F - \|\matr{U}\matr{W}_{\matr{U}}\|^2_\Omega) \nonumber \\
&+ \frac{\lambda_\mathrm{u}}{2}\text{Trace}(\matr{U}^T\matr{L}_\mathrm{u}\matr{U}) + \frac{\lambda_\mathrm{w}}{2}\text{Trace}(\matr{W}_{\matr{U}}\matr{L}_\mathrm{w}\matr{W}_{\matr{U}}^T).
\label{eq:f-grassmann}
\end{align}

Since $f(U)$ depends only on the column space of $U$ due to the invariance property demonstrated above, this function is well-defined on the Grassmann manifold $\text{Gr}(m,r)$. 

The optimization problem now becomes, with an abuse of notation discussed in~\cite[\S2]{BouAbs2015}:
\begin{equation}
\min_{U \in \text{Gr}(m,r)} f(U),
\label{eq:gr-problem}
\end{equation}
where we optimize over $r$-dimensional subspaces in $\mathbb{R}^m$, understanding that in practice we represent these subspaces using orthonormal matrices $U \in \text{St}(m,r)$.

In comparison with the original optimization problem~\eqref{eq:gr-objective}, problem~\eqref{eq:gr-problem} can be thought of as more intricate in the sense that the linear feasible set has been replaced by a nonlinear manifold. However, this is mitigated by the fact that formulation~\eqref{eq:gr-problem} enables the application of Riemannian optimization algorithms~\cite{Manopt2014, TowKoeWei2016, Bergmann2022}, which can efficiently navigate the geometry of the Grassmann manifold while respecting its intrinsic structure. Furthermore,~\eqref{eq:gr-problem} has two advantages over~\eqref{eq:gr-objective}: the dimension of the feasible set is reduced---drastically when $m\ll n$---and the Grassmann manifold is compact.

\subsection{Gradient Computation}

Since $W_U$ minimizes $h(U,\cdot)$, the optimality condition gives us $\frac{\partial h}{\partial W}(U,W_U) = 0$. By the envelope theorem, the directional derivative simplifies to:
\begin{equation}
Df(U)[H] = \frac{\partial h}{\partial U}(U,W_U)[H].
\end{equation}

To facilitate differentiation, we rewrite $h(U,W)$ using only Frobenius norms. We define the mask matrix:
\begin{equation}
\Lambda_{ij} = \begin{cases}
\lambda & \text{if } (i,j) \in \Omega \\
0 & \text{otherwise}.
\end{cases}
\end{equation}

Using the identity $\|\matr{U}\matr{W}\|^2_{\overline{\Omega}} = \|\matr{W}\|^2_F - \|\matr{U}\matr{W}\|^2_\Omega$ (valid when $\matr{U}$ has orthonormal columns), we can rewrite:
\begin{align}
h(\matr{U},\matr{W}) = & \frac{1}{2}\|\matr{C} \odot (\matr{U}\matr{W} - \matr{M})\|^2_F + \frac{\lambda^2}{2}\|\matr{W}\|^2_F \nonumber \\
& - \frac{1}{2}\|\Lambda \odot \matr{U}\matr{W}\|^2_F + \frac{\lambda_\mathrm{u}}{2}\text{Trace}(\matr{U}^T\matr{L}_\mathrm{u}\matr{U}) + \frac{\lambda_\mathrm{w}}{2}\text{Trace}(\matr{W}\matr{L}_\mathrm{w}\matr{W}^T).
\end{align}

This form simplifies differentiation, as for differentiable mappings $g$: 
\begin{equation}
D\left(X \mapsto \frac{1}{2}\|g(X)\|^2_F\right)(X)[H] = \langle Dg(X)[H], g(X) \rangle.
\end{equation}

Defining $\hat{\matr{C}} = \matr{C}^{(2)} - \boldsymbol{\Lambda}^{(2)}$ (where $(\cdot)^{(2)}$ denotes element-wise squaring), we obtain:
\begin{equation}
Df(U)[H] = \langle H, [\hat{C} \odot (UW_U - \matr{M}_\Omega) - \lambda^2 \matr{M}_\Omega]W_U^T + \lambda_\mathrm{u} \matr{L}_\mathrm{u} U \rangle.
\end{equation}

Define the residue matrix:
\begin{equation}
R_U = \hat{C} \odot (UW_U - \matr{M}_\Omega) - \lambda^2 \matr{M}_\Omega.
\label{eq:residue-matrix}
\end{equation}

This gives us the Euclidean gradient:
\begin{equation}
\nabla f(U) = R_U W_U^T + \lambda_\mathrm{u} \matr{L}_\mathrm{u} U.
\end{equation}

In view of~\cite[\S 2]{BouAbs2015}, and with the same abuse of notation and language, the Riemannian gradient on the Grassmannian is obtained by projecting onto the tangent space:
\begin{equation}
\text{grad } f(U) = (I - UU^T)\nabla f(U) = (I - UU^T)(R_U W_U^T + \lambda_\mathrm{u} \matr{L}_\mathrm{u} U).
\end{equation}

\subsection{Hessian Computation}

We now differentiate the gradient expression to obtain the Hessian of $f$. We define the vector field $\bar{F}: \mathbb{R}_*^{m \times r} \rightarrow \mathbb{R}^{m \times r}$:
\begin{equation}
\bar{F}(U) = (I - UU^T)(R_U W_U^T + \lambda_\mathrm{u} \matr{L}_\mathrm{u} U).
\end{equation}

Distributing the projector, we can rewrite this as:
\begin{equation}
\bar{F}(U) = R_U W_U^T + \lambda_\mathrm{u} \matr{L}_\mathrm{u} U - UU^T R_U W_U^T - \lambda_\mathrm{u} UU^T \matr{L}_\mathrm{u} U.
\end{equation}

The Riemannian Hessian of $f$ on the Grassmannian is represented by:
\begin{equation}
\text{Hess } f(U)[H] = (I - UU^T)D\bar{F}(U)[H],
\end{equation}
$\forall H \in \mathbb{R}^{m\times r}$ with $U^T H = 0$; see~\cite[\S 2]{BouAbs2015}.

Computing the differential of $\bar{F}$ term by term yields:
\begin{align}
D\bar{F}(U)[H] &= [\hat{C} \odot (HW_U + UW_{U,H})]W_U^T + R_U W_{U,H}^T + \lambda_\mathrm{u} \matr{L}_\mathrm{u} H \nonumber \\
&\quad- (HU^T + UH^T) R_U W_U^T - UU^T [\hat{C} \odot (HW_U + UW_{U,H})]W_U^T\nonumber \\
&\quad- UU^T R_U W_{U,H}^T - \lambda_\mathrm{u} (HU^T + UH^T) \matr{L}_\mathrm{u} U - \lambda_\mathrm{u} UU^T \matr{L}_\mathrm{u} H.
\end{align}

Applying the projector $(I - UU^T)$ and using $(I - UU^T)U = 0$ to eliminate terms, the final expression for the Hessian becomes:
\begin{align}
\text{Hess } f(U)[H] = &(I - UU^T)([\hat{C} \odot (HW_U + UW_{U,H})]W_U^T + R_U W_{U,H}^T + \lambda_\mathrm{u} \matr{L}_\mathrm{u} H) \nonumber \\
&- HU^T(R_U W_U^T + \lambda_\mathrm{u} \matr{L}_\mathrm{u} U).
\end{align}

The computation requires $W_{U,H}$, the differential of the mapping $U \mapsto W_U$ along $H$, which is provided in the following subsection.

\subsection{Computing $W_U$ and Its Derivative}

We now derive a clean and explicit formulation for computing $W_U$, the optimal right factor given a fixed left factor $U$, as well as its directional derivative $W_{U,H}$. These computations are central to evaluating the cost function $f(U)$ and its Riemannian Hessian. Following the approach of~\cite{BouAbs2015}, we use vectorization and Kronecker product properties to derive closed-form expressions.

For a fixed $U \in \text{St}(m,r)$, computing $W_U$ amounts to solving the least-squares problem~\eqref{eq:wu-definition}:
\begin{equation}
h(U, W) = \frac{1}{2} \|C \odot (UW - \matr{M}_\Omega)\|_F^2 + \frac{\lambda^2}{2} \|W\|^2_F - \frac{\lambda^2}{2} \|UW\|^2_\Omega + \frac{\lambda_\mathrm{w}}{2} \text{Tr}(W \matr{L}_\mathrm{w} W^T).
\end{equation}

To facilitate vectorization, we use the following notation. Let $w = \text{vec}(W) \in \mathbb{R}^{rn}$ be the vectorization of $W$.

Following~\cite{BouAbs2015}, we define the matrices $S \in \mathbb{R}^{k \times mn}$ and operators that will allow us to express the objective in standard quadratic form:
\begin{equation}
S = I_\Omega \text{diag}(\text{vec}(C)),
\end{equation}
where $I_\Omega \in \mathbb{R}^{k \times mn}$ is the sampling operator such that $I_\Omega \text{vec}(M)$ extracts the entries of $M$ corresponding to indices in $\Omega$.

We can now express the objective function $h(U,W)$ in terms of $w$:

\allowdisplaybreaks
\begin{align*}
    h(U,W) &= \frac{1}{2}\sqfrobnorm[\Omega]{C\odot (UW-M_\Omega)} + \frac{\lambda^2}{2}\sqfrobnorm{W} - \frac{\lambda^2}{2}\sqfrobnorm[\Omega]{UW} + \frac{\lambda_u^2}{2}\sqnorm[\calD,u]{U} \\
    &\quad+ \frac{\lambda_w^2}{2}\sqnorm[\calD,w]{W\tr}\\
    &= \frac{1}{2}\sqnorm{S \veccc{UW} - \veccc[\Omega]{C\odot M_\Omega}}\\
    &\quad + \frac{\lambda^2}{2}\sqnorm{\veccc{W}} - \frac{\lambda^2}{2}\sqnorm{\veccc[\Omega]{UW}} + \frac{\lambda_u^2}{2}\sqnorm[\calD,u]{U} + \frac{\lambda_w^2}{2}\sqnorm[\calD,w]{W\tr}\\
    &=\frac{1}{2}\sqnorm{S(I_n \otimes U)\veccc{W}-\veccc[\Omega]{C\odot M_\Omega}} + \frac{1}{2}\sqnorm{\lambda I_{rn}\veccc{W}} \\
    &\quad- \frac{1}{2}\sqnorm{\lambda I_\Omega (I_n \otimes U)\veccc{W}} + \frac{\lambda_u^2}{2}\sqnorm[\calD,u]{U} + \frac{\lambda_w^2}{2}\sqnorm[\calD,w]{W\tr}\\
    &= \frac{1}{2}\sqnorm{\begin{bmatrix} S(I_n \otimes U) \\ \lambda I_{rn} \end{bmatrix} \veccc{W} - \begin{bmatrix} \veccc[\Omega]{C\odot M_\Omega}\\0_{rn} \end{bmatrix}} \\
    &\quad- \frac{1}{2}\sqnorm{\lambda I_\Omega (I_n \otimes U)\veccc{W}}+\frac{\lambda_w^2}{2}\Trace(WL_wW\tr) +\frac{\lambda_u^2}{2}\sqnorm[\calD,u]{U}\\
    &= \frac{1}{2}\sqnorm{\begin{bmatrix} S(I_n \otimes U) \\ \lambda I_{rn} \end{bmatrix} \veccc{W} - \begin{bmatrix} \veccc[\Omega]{C\odot M_\Omega}\\0_{rn} \end{bmatrix}} +\frac{\lambda_u^2}{2}\sqnorm[\calD,u]{U} \\
    &\quad- \frac{1}{2}\sqnorm{\lambda I_\Omega (I_n \otimes U)\veccc{W}}+ \frac{\lambda_w^2}{2}\veccc{W}\tr(L_w\otimes I_r)\veccc{W}\\
    &=\frac{1}{2}\sqnorm{A_1w - b_1}-\frac{1}{2}\sqnorm{A_2w}+\frac{\lambda_w^2}{2}w\tr(L_w\otimes I_r)w+\frac{\lambda_u^2}{2}\sqnorm[\calD,u]{U}\\
    &=\frac{1}{2}w\tr(A_1\tr A_1 - A_2\tr A_2 + \lambda_w^2(L_w\otimes I_r))w - b_1\tr A_1 w + \frac{1}{2}\sqnorm{b_1} + \frac{\lambda_u^2}{2}\sqnorm[\calD,u]{U}\\
    &=\frac{1}{2}w\tr A w - b\tr w + \text{const},
\end{align*}
where:
\begin{align}
A &= (I_n \otimes U^T) S^T S (I_n \otimes U) + \lambda^2 I_{rn} - \lambda^2 (I_n \otimes U^T) I_\Omega^T I_\Omega (I_n \otimes U) + \lambda_\mathrm{w}^2 (\matr{L}_\mathrm{w} \otimes I_r). \\
b &= (I_n \otimes U^T) S^T \text{vec}_\Omega(C \odot \matr{M}_\Omega).
\end{align}

Define $\hat{C} = C^{(2)} - \Lambda^{(2)}$ where $(\cdot)^{(2)}$ denotes element-wise squaring. Then:
\begin{align}
A &= (I_n \otimes U^T) \text{diag}(\text{vec}(\hat{C})) (I_n \otimes U) + \lambda^2 I_{rn} + \lambda_\mathrm{w}^2 (\matr{L}_\mathrm{w} \otimes I_r). \\
b &= (I_n \otimes U^T) \text{vec}(C^{(2)} \odot \matr{M}_\Omega).
\end{align}

The solution is given by:
\begin{equation}
\text{vec}(W_U) = A^{-1} b.
\label{eq:wu-solution}
\end{equation}

The key difference from the original RTRMC formulation is the additional term $\lambda_\mathrm{w} (\matr{L}_\mathrm{w} \otimes I_r)$ in the matrix $A$, which introduces coupling between all columns of $W$ via the graph Laplacian $\matr{L}_\mathrm{w}$. This represents a significant departure from the block-diagonal structure in the original RTRMC, where each column of $W$ could be computed independently.

Let us compute the directional derivative $W_{U,H}$ of the mapping $U \mapsto W_U$ along a direction $H$:
\begin{equation}
\text{vec}(W_{U,H}) = D(U \mapsto \text{vec}(W_U))(U)[H].
\end{equation}

Using the formula for the differential of the inverse of a matrix and the product rule:
\begin{align}
\text{vec}(W_{U,H}) &= -A^{-1} \cdot D(U \mapsto A)(U)[H] \cdot \text{vec}(W_U) + A^{-1} \cdot D(U \mapsto b)(U)[H].
\end{align}

Computing the differentials:
\begin{equation}
D(U \mapsto b)(U)[H] = (I_n \otimes H^T) \text{vec}(\hat{C} \odot \matr{M}_\Omega) = \text{vec}(H^T (\hat{C} \odot \matr{M}_\Omega)).
\end{equation}
\begin{align}
D(U \mapsto A)(U)[H] &= (I_n \otimes H^T) \text{diag}(\text{vec}(\hat{C})) (I_n \otimes U) \\&\quad+ (I_n \otimes U^T) \text{diag}(\text{vec}(\hat{C})) (I_n \otimes H). \nonumber
\end{align}

Let $B = \text{diag}(\text{vec}(\hat{C}))$. Then:
\begin{align}
\text{vec}(W_{U,H}) &= -A^{-1} [(I_n \otimes H^T) B (I_n \otimes U) + (I_n \otimes U^T) B (I_n \otimes H)] \text{vec}(W_U)\\
&\quad + A^{-1} \text{vec}(H^T (\hat{C} \odot \matr{M}_\Omega))  \nonumber\\
&= -A^{-1} [(I_n \otimes H^T) \text{vec}(\hat{C} \odot UW_U) + (I_n \otimes U^T) \text{vec}(\hat{C} \odot HW_U))]  \nonumber\\
&\quad + A^{-1} \text{vec}(H^T (\hat{C} \odot \matr{M}_\Omega))  \nonumber\\
&= -A^{-1} [\text{vec}(H^T (\hat{C} \odot UW_U)) + \text{vec}(U^T (\hat{C} \odot HW_U))]  \nonumber\\
&\quad + A^{-1} \text{vec}(H^T (\hat{C} \odot \matr{M}_\Omega)).  \nonumber
\end{align}

Recalling the definition of the residue matrix $R_U = \hat{C} \odot (UW_U - \matr{M}_\Omega)$ from equation \eqref{eq:residue-matrix}, we obtain the compact expression:
\begin{equation}
\text{vec}(W_{U,H}) = -A^{-1} \left[ \text{vec}(H^T R_U) + \text{vec}(U^T (\hat{C} \odot HW_U)) \right].
\label{eq:wu-h-final}
\end{equation}

\subsection{Initialization Strategy}

The GR-RTRMC algorithm requires an initial point on the Grassmann manifold $\text{Gr}(m,r)$, represented by an orthonormal matrix $U_0 \in \text{St}(m,r)$. Unlike factorization-based approaches that require initialization of both matrices, our formulation only requires initializing the left matrix $U$, since the right matrix $W$ is computed optimally via the least-squares solution \eqref{eq:wu-solution}.

We initialize $U_0$ using the $r$ dominant left singular vectors of the observed data matrix $\matr{M}_\Omega$:
\begin{equation}
U_0 = \text{SVD}_r(\matr{M}_\Omega)_{\text{left}},
\end{equation}
where $\text{SVD}_r(\cdot)_{\text{left}}$ denotes the matrix of the $r$ leading left singular vectors. Since $\matr{M}_\Omega$ is sparse with $k = |\Omega|$ nonzero entries, this computation can be performed efficiently using the PROPACK sparse SVD algorithm~\cite{PROPACK}, with computational cost $O(kr + r^3)$.

\section{Numerical Experiments}
\label{sec:experiments}
All experiments are performed on a desktop with 4.5 GHz AMD Ryzen 9 7950X with 64GB of RAM running Julia 1.10.4. Reported computational times exclude the initial compilation overhead by discarding the first execution.

We conduct experiments on two types of datasets: weather data (Belgium and France) and MovieLens.

For the meteorological datasets, we follow the experimental protocol (data splits, masking scenarios, and evaluation metrics) and station-graph construction described in~\cite{loucheur2023graphbased}. The data matrix has rows corresponding to weather stations and columns to time steps, so that the row graph encodes spatial proximity between stations. The spatial graph connecting weather stations is built using inverse distance weighting with altitude corrections, among other approaches explored in that work; see also~\cite{graph_form} for graph construction in matrix completion. We consider two missing data scenarios: \emph{Block}, where contiguous time intervals are missing at selected stations, and \emph{Spread}, where missing entries are distributed more uniformly across time and stations. For all methods, hyperparameters (including the rank $r$ and regularization parameters) are selected via cross-validation on a validation set, and the reported results are computed on a separate test set.

\subsection{Methods used for comparison}
For the weather datasets, the following methods were employed for matrix completion:
\begin{enumerate}
    \item \textbf{IDW (Inverse Distance Weighting)}~\cite{IDW, ALEJOSANCHEZ2025103455}\\
    A simple interpolation method that estimates missing values by calculating a weighted average of nearby observations. The weights decrease as the distance from the missing value increases.

    \item \textbf{PCA (Principal Component Analysis)}~\cite{PCA, ALEJOSANCHEZ2025103455} \\
    A dimensionality reduction technique that imputes missing values by projecting the data onto a lower-dimensional space defined by its principal components.

    \item \textbf{MissForest}~\cite{missforest} \\
    A machine learning-based imputation method that uses random forests. Missing values are predicted iteratively based on the relationships between observed variables.

    \item \textbf{LMaFiT (Low-rank Matrix Fitting)}~\cite{lmafit} \\
    A matrix factorization approach that models the observed data as a low-rank matrix. Missing values are estimated through iterative optimization.

    \item \textbf{SoftImpute}~\cite{SoftImpute} \\
    A matrix completion algorithm that employs soft-thresholded singular value decomposition (SVD) to approximate missing entries by minimizing the reconstruction error.

    \item \textbf{RTRMC (Riemannian Trust-Region Matrix Completion)}~\cite{RTRMC_ABSIL_BOUMAL} \\
    A matrix completion technique that operates on the Riemannian manifold of low-rank matrices, employing trust-region optimization to estimate missing data.

    \item \textbf{GRALS (Graph Regularized Alternating Least Squares)}~\cite{grals} \\
    An imputation method that incorporates graph-based regularization with alternating least squares to leverage structural information within the dataset.
\end{enumerate}

\subsection{Weather Data}
Missing data in time series from weather stations can result from a variety of reasons like sensor malfunctions, power failures, communication disruptions, maintenance or removal of misleading values.
Addressing these gaps is however essential for climate and weather analyses, which mostly rely on complete datasets.

Before reporting results, we make explicit how the weather imputation task instantiates the graph-regularized problem~\eqref{eq:gr-objective}. The data matrix $\matr{M}\in\mathbb{R}^{m\times n}$ collects the temperature measurements: its $m$ rows index the weather stations and its $n$ columns index the regularly sampled time steps of the considered time window (for instance $n=1008$ for the Belgian dataset). The confidence matrix $\matr{C}$ is set to one on observed entries and to zero elsewhere, i.e., uniform confidence in the recorded values; the only exception is the storm case study (\Cref{fig:storm_issue_corrected_case}), where $C_{i,j}$ is deliberately lowered over an anomalous time window. Since the rows are the stations, the row Laplacian $\matr{L}_\mathrm{u}\in\mathbb{R}^{m\times m}$ encodes the \emph{spatial} graph between stations, following~\cite{loucheur2023graphbased}. This graph is a $k$-nearest-neighbour graph on the inter-station haversine distance, that is, the surface distance between two points given their latitude and longitude, with inverse-distance edge weights and an altitude-difference threshold. The column Laplacian $\matr{L}_\mathrm{w}\in\mathbb{R}^{n\times n}$ encodes the \emph{temporal} graph, which links nearby time steps so as to promote temporal smoothness. The model therefore exposes parameters to be chosen: the rank $r$, the three regularization weights $\lambda$, $\lambda_\mathrm{u}$, $\lambda_\mathrm{w}$ of~\eqref{eq:gr-objective}, and the graph construction method and its parameters.

\subsubsection*{In Belgium}
\begin{figure}[ht!]
    \centerline{\includegraphics[scale=0.4]{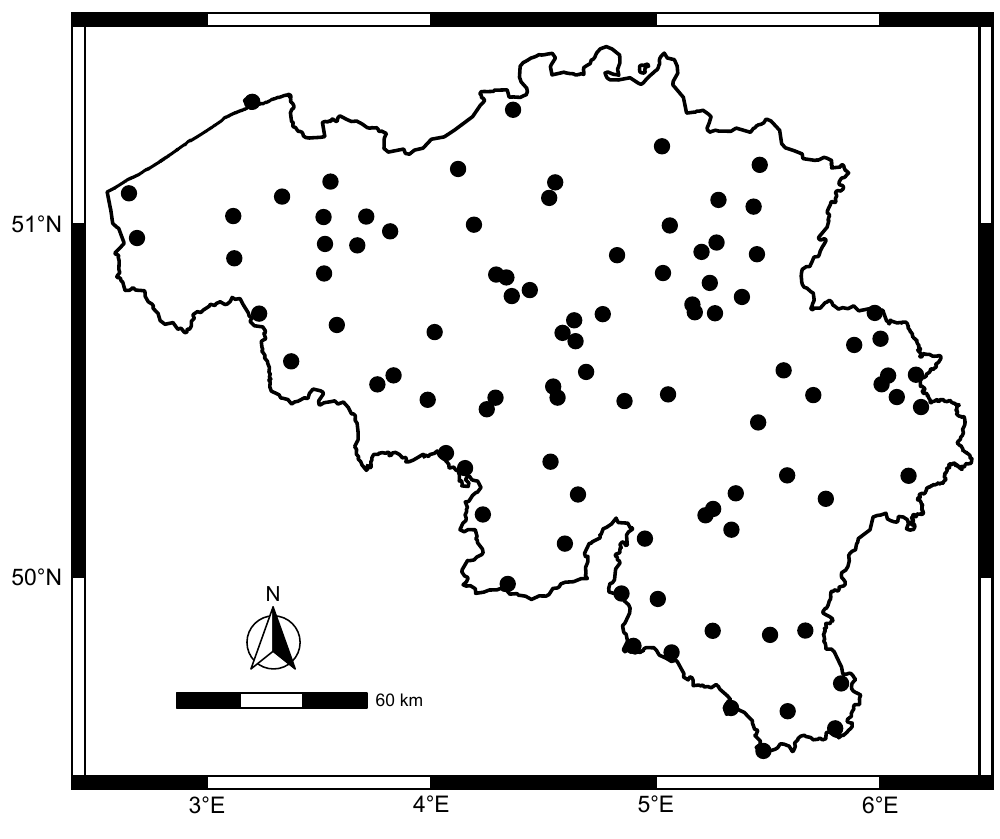}}
    \caption{Position of the 96 weather stations in Belgium.}
    \label{fig:IRM_stations}
\end{figure}

For our experiments, we used a dataset provided by the Royal Meteorological Institute of Belgium (RMI), which includes temperature measurements recorded every ten minutes across 96 weather stations in Belgium as represented in~\cref{fig:IRM_stations}. The analysis covers the period from January 1, 2020 to December 31, 2023. Results are summarized in \Cref{table:irm_results_all}.

\begin{table}[h!]
    \centering
	\caption{Average RMSE (in $^\circ$C) and average computational time (in seconds) on the test set for every method considered. Results for the two types of missing data generation on the Belgian weather dataset (RMI)}
	\label{table:irm_results_all}
	\vskip 0.10in
	\begin{small}
	\begin{sc}
	\begin{tabular}{lcccl}\toprule
		& \multicolumn{2}{c}{Block} & \multicolumn{2}{c}{Spread}
		\\\cmidrule(lr){2-3}\cmidrule(lr){4-5}
			Methods	& RMSE  & Time	& RMSE	& Time\\\midrule
		IDW 		&0.64	&0.14	&0.57	&0.19	\\
		PCA 		&0.61	&0.25	&0.68	&0.35	\\
        MissForest & 0.59 & 10.1 & 0.56 & 9.6 \\\midrule
		LMaFiT 		&0.81	&4.9	&0.76	&4.6	\\
		SoftImpute 	&0.53	&3.1	&\textbf{0.41}&2.9	\\
		RTRMC 		&0.49	&3.4	&0.46	&3.7	\\\midrule
		GRALS 		&0.53	&8.0	&0.63	&12.5	\\
		GR-RTRMC   	&\textbf{0.45}	& 8.6	&0.43	&	9.6	\\\bottomrule
	\end{tabular}
	\end{sc}
	\end{small}
	\vskip -0.15in
\end{table}

\begin{figure}[ht!]
    \centering
    \includegraphics[width=1\textwidth]{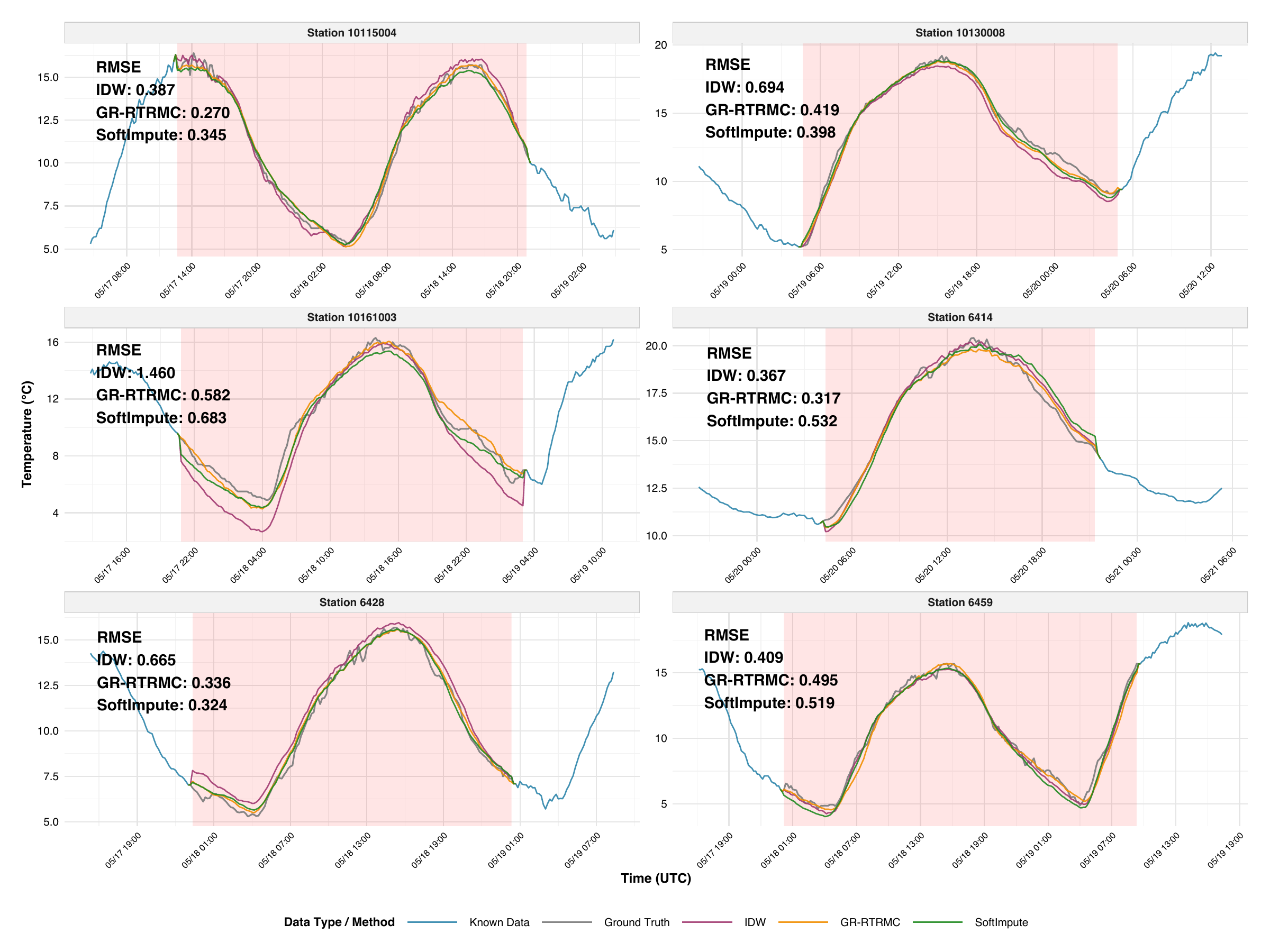}
    \caption{Imputation comparison results on 6 Belgian stations showing different completion methods on the test set}
    \label{fig:imputation_comparison}
\end{figure}

\begin{figure}[ht!]
    \centering
    \begin{subfigure}[t]{0.75\textwidth}
        \centering
        \includegraphics[scale=0.3]{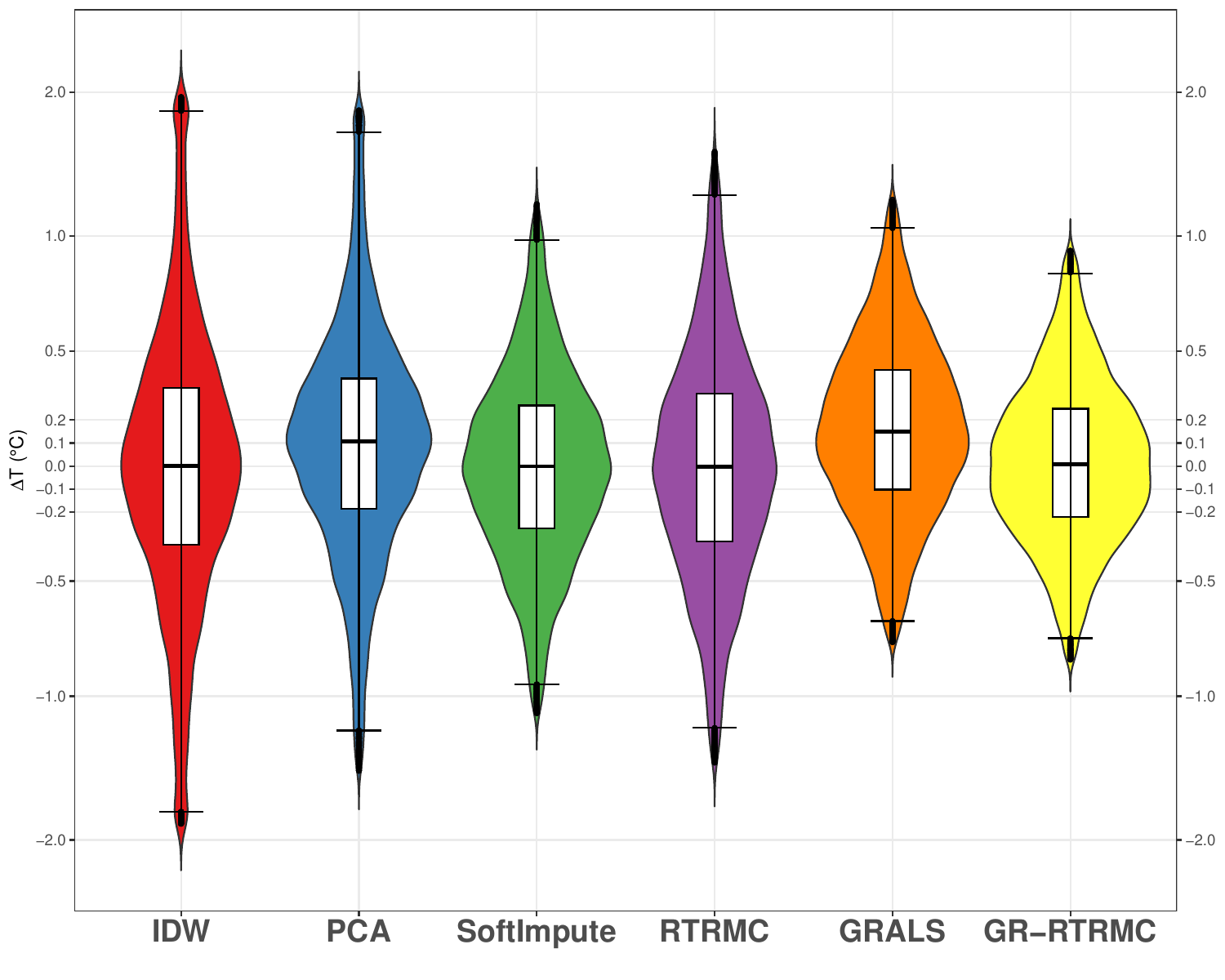}
    \end{subfigure}
    \begin{subfigure}[t]{0.2\textwidth}
        \centering
        \includegraphics[scale=0.5]{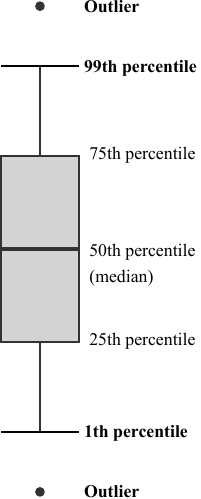}
    \end{subfigure}
    \caption{Violin plot of completion errors (in $^\circ$C) for every method considered on the Belgian test set in the Block scenario}
    \label{fig:violin}
\end{figure}

GR-RTRMC achieves the best overall performance on the Belgian dataset, with the lowest RMSE of 0.45°C in the Block scenario. In the Spread scenario, SoftImpute achieves the lowest RMSE (0.41°C), while GR-RTRMC remains competitive (0.43°C). \Cref{fig:imputation_comparison} shows the completion quality across multiple test stations, revealing that while GR-RTRMC has the lowest average RMSE, it is not systematically superior at every individual station. The violin plot (\Cref{fig:violin}) illustrates the distribution of completion errors across the test set, showing the variability in method performance.

\subsubsection*{In France} 
\begin{figure}[ht!]
    \centerline{\includegraphics[scale=0.4]{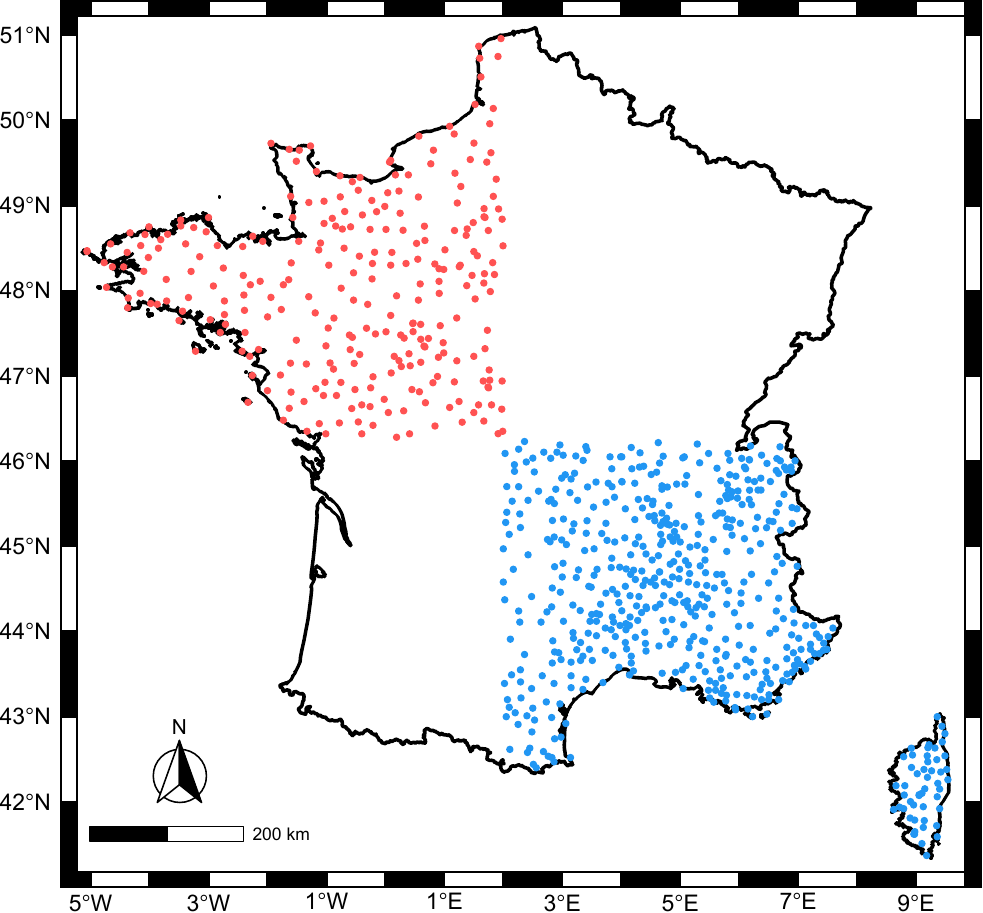}}
    \caption{Position of the weather stations in the northwest (NW) and southeast (SE) regions of France.}
    \label{fig:MeteoFrance_stations}
\end{figure}

For our experiments, we used a dataset provided by Météo-France,\footnote{\url{https://meteonet.umr-cnrm.fr/}} which includes temperature measurements recorded every six minutes across 272 weather stations in the northwest (NW) region and 545 stations in the southeast (SE) region of France. The data covers the three-year period from 2016 to 2018. The two regions are represented as rectangular areas in~\cref{fig:MeteoFrance_stations}.

\begin{table}[h!]
    \vspace{-0.3cm}
    \centering
    \caption{Average RMSE (in $^\circ$C) and average computational time (in seconds) on the test set for every method considered. Results for the two types of missing data generation on the French weather dataset (M\'{e}t\'{e}o-France)}
    \label{table:meteofrance_results_all}
    \vskip 0.10in
    \begin{small}
    \begin{sc}
    \begin{tabular}{lcccl}\toprule
            & \multicolumn{2}{c}{Block} & \multicolumn{2}{c}{Spread}
            \\\cmidrule(lr){2-3}\cmidrule(lr){4-5}
                Methods	& RMSE  & Time	& RMSE	& Time\\\midrule
            IDW 		&0.80	&0.21	&0.74	&0.19	\\
            PCA 		&0.77	&0.24	&0.67	&1.1	\\
            MissForest  &0.70   &14.3   &0.56  &13.8   \\\midrule
            LMaFiT 		&0.94	&9.7	&1.01	&9.3	\\
            SoftImpute 	&0.59	&7.3	&0.61	&6.7	\\
            RTRMC 		&0.67	&8.2	&0.58	&4.9	\\\midrule
            GRALS 		&0.62	&20.1	&0.68	&19.3	\\
            GR-RTRMC    	&\textbf{0.57}	&10.1	&\textbf{0.54}	&13.1
            \\\bottomrule
        \end{tabular}
    \end{sc}
    \end{small}
    \vskip -0.15in
\end{table}

On the France dataset, GR-RTRMC demonstrates consistent performance with an RMSE of 0.57°C (Block) and 0.54°C (Spread), outperforming most baselines including GRALS (\Cref{table:meteofrance_results_all}). While SoftImpute shows interesting results, particularly in the Block scenario, GR-RTRMC achieves the best overall performance in the Spread scenario, confirming the effectiveness of graph regularization. The advantage of graph-based methods becomes more pronounced at larger scales, as the French dataset with its higher number of stations provides richer spatial relationships for the graph regularization to exploit.


\subsection{MovieLens}
\label{sec:movielens_exp}
The MovieLens 100K dataset~\cite{10.1145/2827872} is a well-established resource in the realm of recommender systems, containing 100,000 ratings from 943 users for 1682 movies. The data was collected during the seven months from September 19, 1997 to April 22, 1998. Each user in the dataset has rated at least 20 movies, ensuring sufficient user-item interactions for meaningful analysis.
The MovieLens 1M dataset will also be used, which contains 1,000,029 ratings from 6040 users for 3952 movies. In both cases, the data matrix has rows corresponding to users and columns to movies, with the row graph encoding user similarity and the column graph encoding movie similarity.

For the MovieLens experiments, we employ the same matrix completion methods as for the weather datasets (LMaFiT, SoftImpute, RTRMC, GRALS, GR-RTRMC), but replace IDW and PCA with User-Based KNN, which is more appropriate for collaborative filtering.

We construct a user–user similarity graph via $k$-nearest neighbors (k-NN) on user rating vectors. Both the neighborhood size $k$ and the similarity function (e.g., Euclidean, cosine, Jaccard, Manhattan) are treated as hyperparameters and selected by validation. In practice, moderate $k$ values and standard similarity metrics yield comparable performance, indicating robustness to these choices.

We apply the same procedure to build a movie–movie (item–item) similarity graph from item rating vectors, with $k$ and the similarity function treated as hyperparameters.

\subsubsection*{Results}
\begin{table}[t!]
    \centering
	\caption{Average RMSE and average computational time (in seconds) on the test set for every method considered. Results for the MovieLens datasets.}
	\label{table:collaborative_results_all}
	\vskip 0.10in
	\begin{small}
	\begin{sc}
	\begin{tabular}{lcccl}\toprule
		& \multicolumn{2}{c}{ML 100K} & \multicolumn{2}{c}{ML 1M}
		\\\cmidrule(lr){2-3}\cmidrule(lr){4-5}
			Methods	& RMSE  & Time	& RMSE	& Time\\\midrule
		User-Based KNN&	0.969&	23.4	&	0.961	&  209.4	\\\midrule
		LMaFiT 		&0.993	&  5.5	&	0.965	&	68.9	\\
		SoftImpute 	&0.973	&  4.6	&	0.932	&	30.1	\\
        MissForest & 0.987 & 3.8 & 0.956 & 24.1\\
		RTRMC 		&0.966	&  4.9	&	0.945	&	39.2	\\\midrule
		GRALS 		&0.951	&  10.4	&	0.981	&	71.2	\\
		GR-RTRMC   	&\textbf{0.942}&  17.3	&	\textbf{0.913}	&	79.5	\\\bottomrule
	\end{tabular}
	\end{sc}
	\end{small}
	\vskip -0.15in
\end{table}

The MovieLens experiments demonstrate the effectiveness of our approach in collaborative filtering scenarios. As shown in \Cref{table:collaborative_results_all}, GR-RTRMC achieves the best performance on both datasets. The method obtains RMSE values of 0.942 for MovieLens 100K and 0.913 for MovieLens 1M, outperforming all baseline approaches including GRALS and traditional collaborative filtering methods like User-Based KNN.

\subsection{Computational Complexity}

While GR-RTRMC achieves superior accuracy across nearly all datasets, this performance comes at the cost of increased computational time. As shown in the timing results, GR-RTRMC consistently requires 2-3 times longer computation time than RTRMC due to the additional graph regularization.

For instance, on the Belgian dataset, GR-RTRMC requires 8.6-9.6 seconds compared to RTRMC's 3.4-3.7 seconds. Similar patterns are observed across the French and MovieLens datasets. Despite this computational overhead, the method remains practical for real-world applications where accuracy improvements justify the additional computation time.

\subsection{Case analysis}

When conducting the numerical experiments with the Belgian weather data, we observed isolated cases of abnormally high RMSE at a few stations that appeared only in exceptional runs. We analysed in more detail two of such cases.

\subsubsection*{Case 1: Storm issue}

\begin{figure}[h!]
    \centering
    \includegraphics[width=0.6\linewidth]{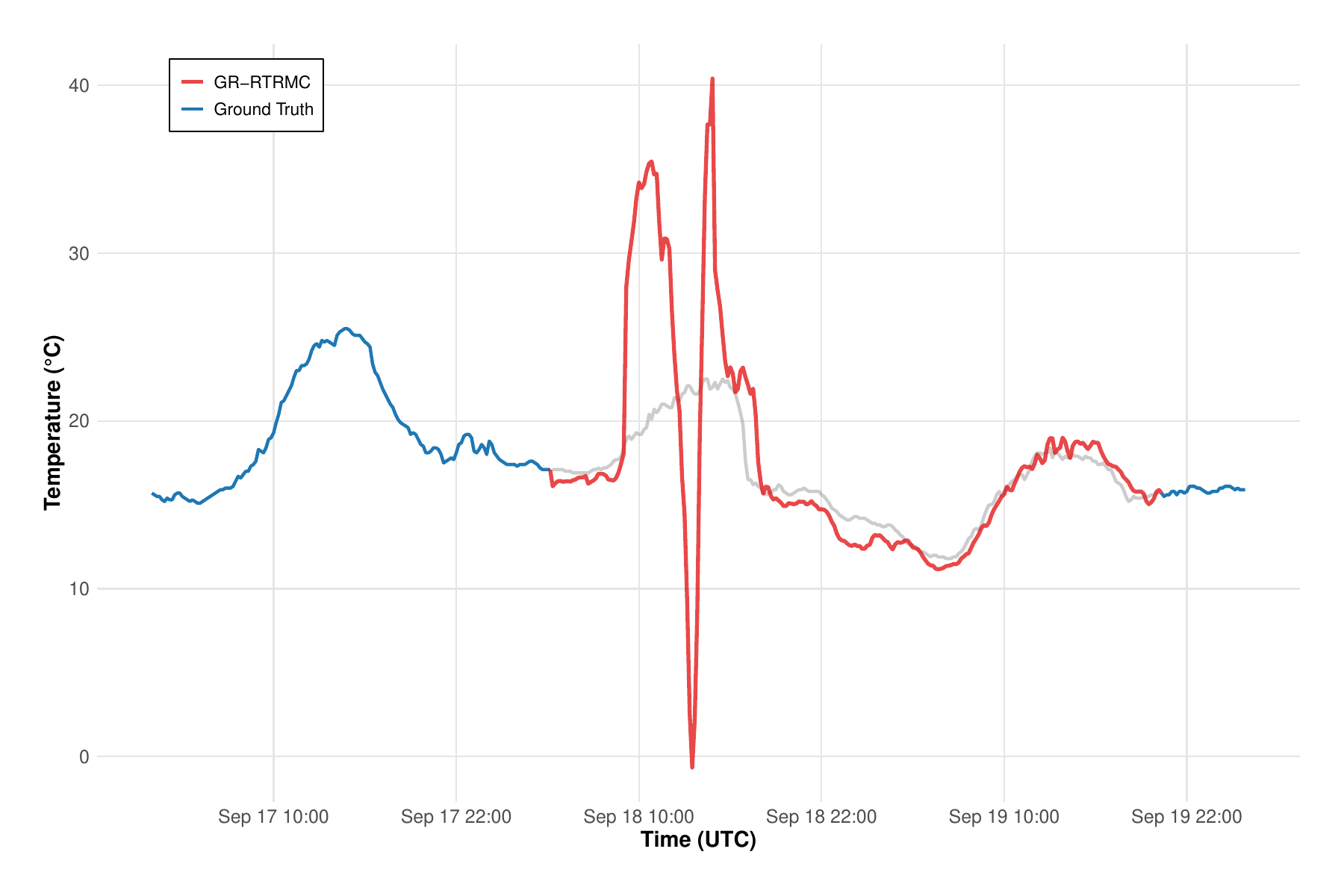}
    \caption{Temperature imputation artifacts during storm event: spurious oscillations in reconstructed temperature data at a Belgian weather station using GR-RTRMC}
    \label{fig:storm_issue_bad_case}
\end{figure}

In one particular imputation week, the reconstructed temperature at some stations contained implausible, narrow oscillations. An illustrative example of this artifact is shown in \Cref{fig:storm_issue_bad_case}.

\begin{figure}[h!]
    \centering
    \begin{subfigure}[t]{0.24\textwidth}
        \centering
        \includegraphics[width=\linewidth]{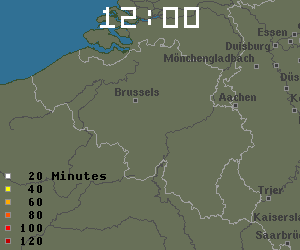}
    \end{subfigure}
    \hfill
    \begin{subfigure}[t]{0.24\textwidth}
        \centering
        \includegraphics[width=\linewidth]{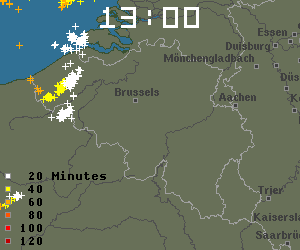}
    \end{subfigure}
    \hfill
    \begin{subfigure}[t]{0.24\textwidth}
        \centering
        \includegraphics[width=\linewidth]{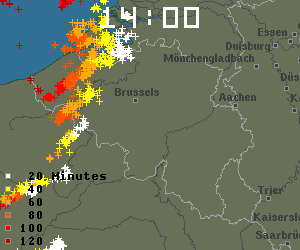}
    \end{subfigure}
    \hfill
    \begin{subfigure}[t]{0.24\textwidth}
        \centering
        \includegraphics[width=\linewidth]{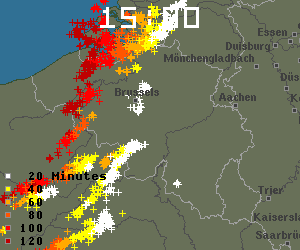}
    \end{subfigure}

    \vspace{0.3cm}

    \begin{subfigure}[t]{0.24\textwidth}
        \centering
        \includegraphics[width=\linewidth]{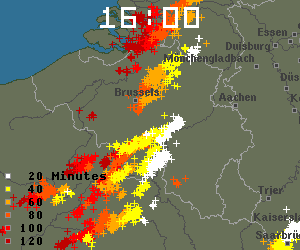}
    \end{subfigure}
    \hfill
    \begin{subfigure}[t]{0.24\textwidth}
        \centering
        \includegraphics[width=\linewidth]{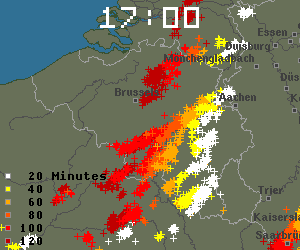}
    \end{subfigure}
    \hfill
    \begin{subfigure}[t]{0.24\textwidth}
        \centering
        \includegraphics[width=\linewidth]{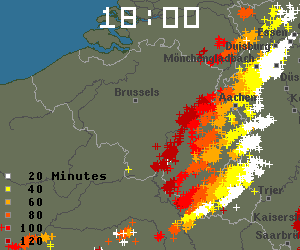}
    \end{subfigure}
    \hfill
    \begin{subfigure}[t]{0.24\textwidth}
        \centering
        \includegraphics[width=\linewidth]{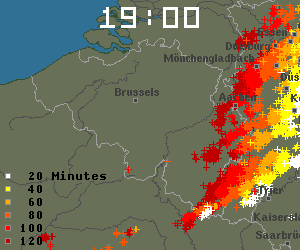}
    \end{subfigure}
    \caption{Storm progression on September 18, 2023, as observed by lightning detection systems, hourly time step from 12:00 to 19:00 UTC}
    \label{fig:storm_progression}
\end{figure}

\begin{figure}[h!]
    \centering
    \begin{subfigure}[c]{0.48\textwidth}
        \centering
        \includegraphics[width=\linewidth]{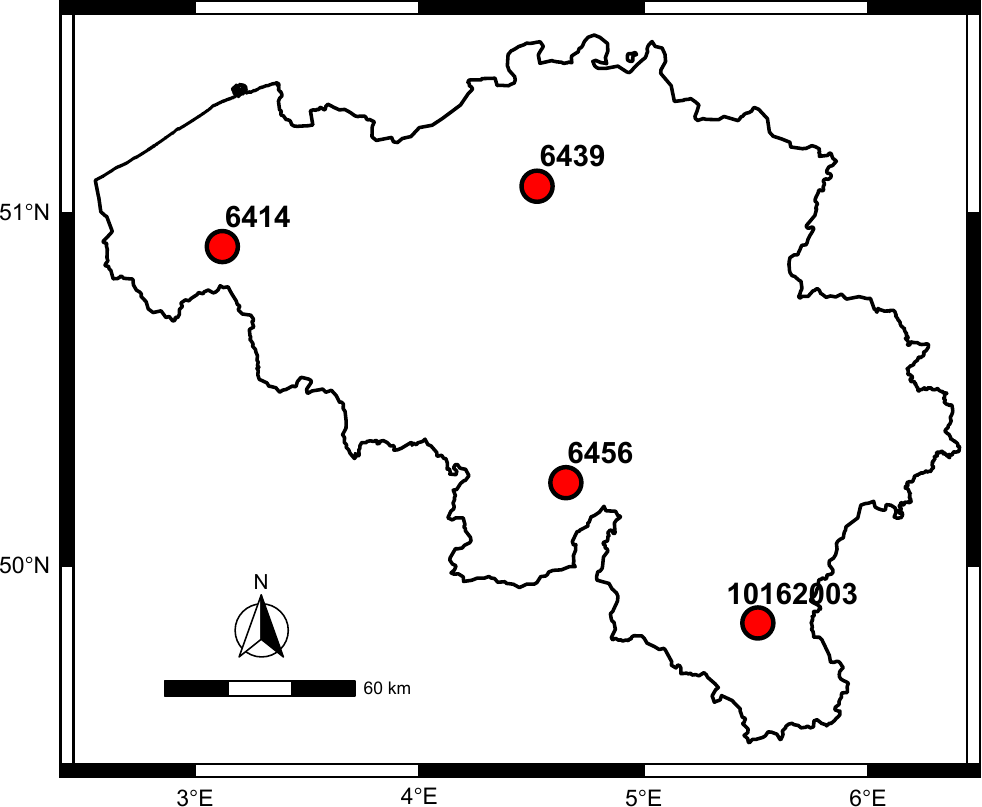}
        \caption{Selected stations map}
        \label{fig:selected_stations_map}
    \end{subfigure}
    \hfill
    \begin{subfigure}[c]{0.48\textwidth}
        \centering
        \includegraphics[width=\linewidth]{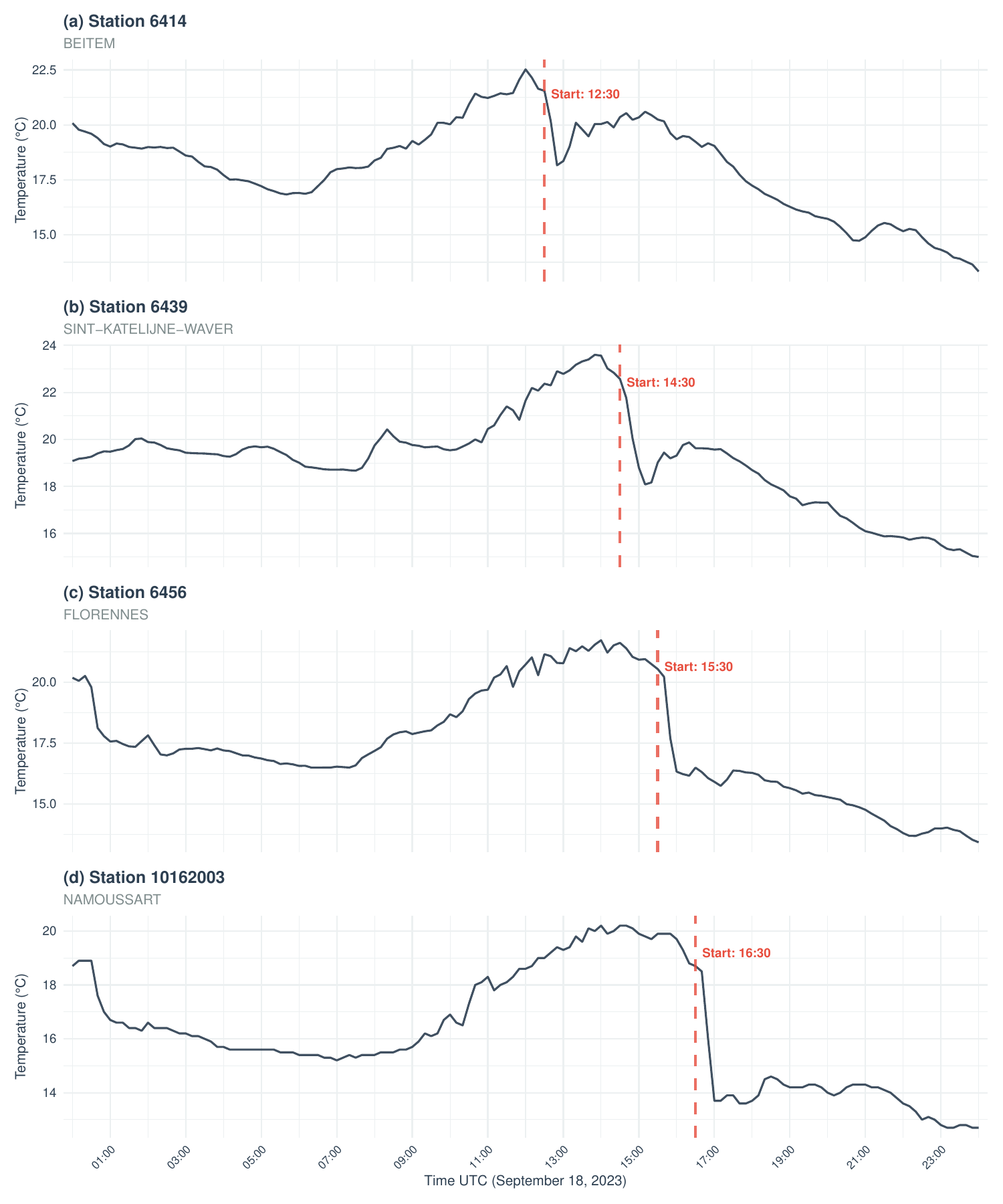}
        \caption{Selected stations plot}
        \label{fig:temperature_analysis_subplots}
    \end{subfigure}
    \caption{Map of the four selected stations (left) and time series of temperatures recorded on the day of the event (right), showing sudden drops in temperature offset by several hours.}
    \label{fig:storm_issue_comparison}
\end{figure}

To understand why this week produced oscillations at multiple stations, we analyzed hourly radar overlays and identified an exceptional convective front crossing Belgium (\Cref{fig:storm_progression}). Its passage time was spatially heterogeneous, with stations tens to hundreds of kilometers apart affected at different hours (see \Cref{fig:selected_stations_map,fig:temperature_analysis_subplots}).

The station map and time series clearly show this heterogeneity: four stations across Belgium all record a sharp temperature drop, but this occurs at different times, staggered by roughly 1–2 hours (\Cref{fig:selected_stations_map,fig:temperature_analysis_subplots}). A low-rank model that assumes time-synchronous behavior, especially when further constrained by spatial graph regularization, cannot capture these temporary phase shifts. When forced to reconcile misaligned signals, the method produces the oscillation observed during the storm.

\begin{figure}[h!]
    \centering
    \includegraphics[width=0.6\linewidth]{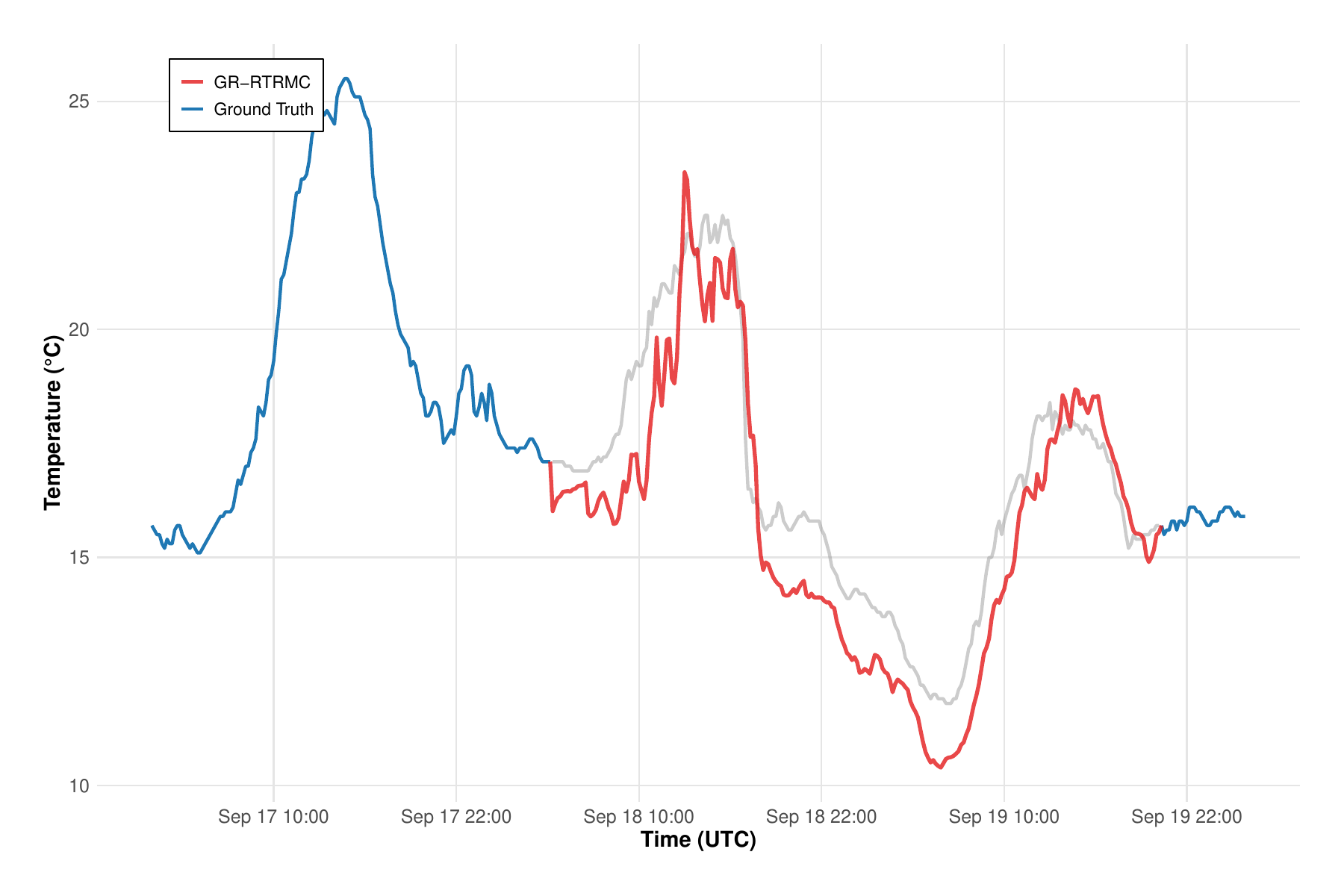}
    \caption{Corrected case after reducing confidence $C_{i,j}$ during the storm window: the spurious spike disappears and the imputation becomes meteorologically plausible}
    \label{fig:storm_issue_corrected_case}
\end{figure}

Reducing the confidence assigned to known data during the storm window, by lowering $C_{i,j}$ for all $i$ and for $j$ within the global storm period, removed this behavior and restored a plausible imputation, as shown in \Cref{fig:storm_issue_corrected_case}.

\subsubsection*{Case 2: Rain issue}

\begin{figure}
    \centering
    \begin{subfigure}[t]{0.48\textwidth}
        \centering
        \includegraphics[width=\linewidth]{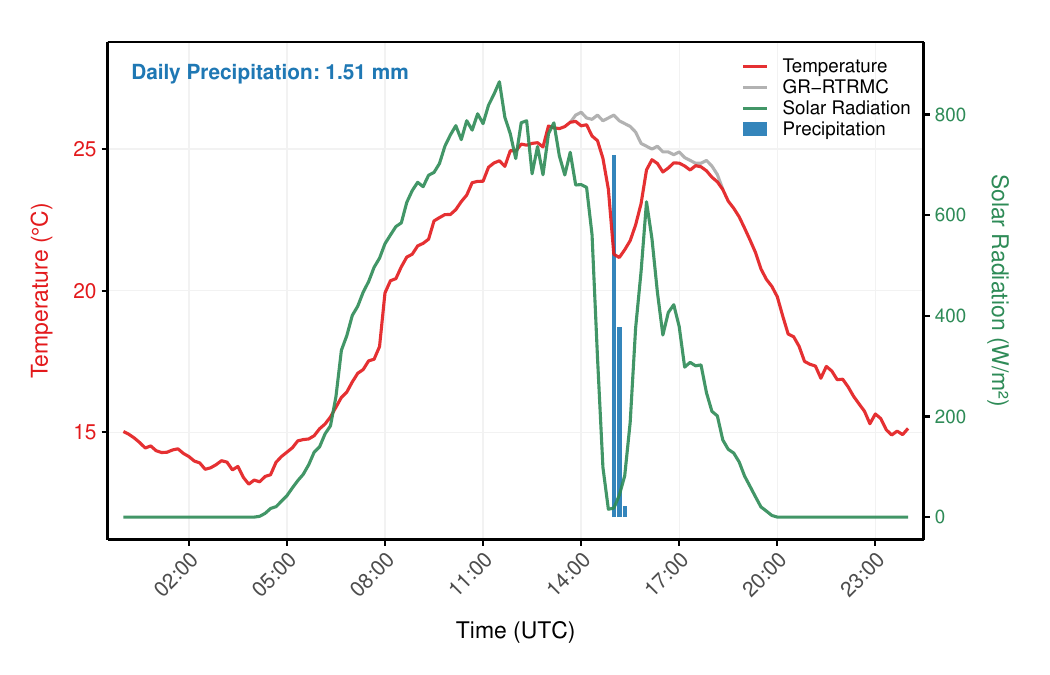}
        \caption{Weather station \texttt{6414}}
    \end{subfigure}
    \vspace{0.5cm}
    \begin{subfigure}[t]{0.48\textwidth}
        \centering
        \includegraphics[width=\linewidth]{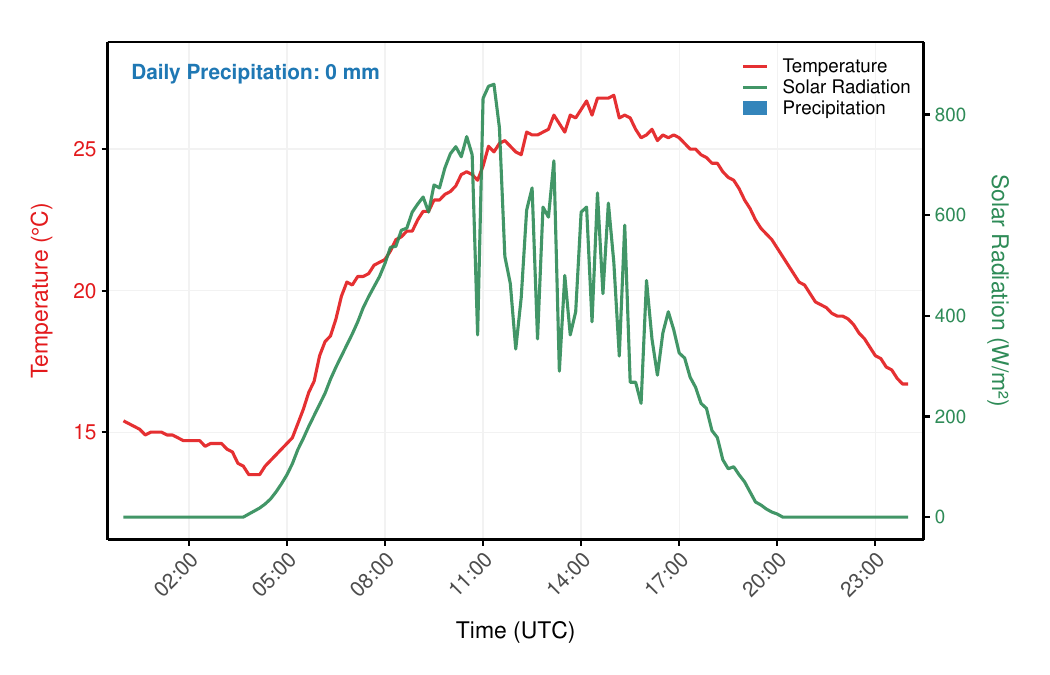}
        \caption{Weather station \texttt{10115001}}
    \end{subfigure}
    \caption{Temperature, solar radiation and precipitation for two closely located weather stations on June 23, 2023. Station \texttt{6414} experiences a sharp, rain-forced cooling between 15:00 and 16:00, whereas station \texttt{10115001}, about 10 km away, remains comparatively unaffected.}
    \label{fig:rain_station_timeseries}
\end{figure}

Rain from scattered thunderstorms creates trouble for accurate imputation of air temperature data. Nearby weather stations often see very different amounts of rainfall, even where they are close together, as illustrated in \Cref{fig:rain_station_timeseries} which shows the temperature, precipitation, solar radiation and predicted temperature using GR-RTRMC for the station \texttt{6414}. In this case, station \texttt{6414} observed heavy rain from 3pm to 4pm on June 23, 2023. The rain caused temperature to drop quickly. However, a weather station about 10 km away (\texttt{10115001}) did not experience any rain during that time. This focused influence makes it tricky to fill in missing data using only temperature information. Nearby locations do not share similar weather conditions at that moment.

\begin{figure}[h!]
    \centering
    \begin{subfigure}[t]{0.48\textwidth}
        \centering
        \includegraphics[width=\linewidth]{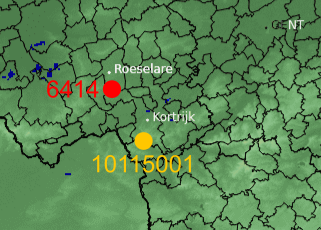}
        \caption{14h00}
        \label{fig:radar_image1}
    \end{subfigure}
    \hfill
    \begin{subfigure}[t]{0.48\textwidth}
        \centering
        \includegraphics[width=\linewidth]{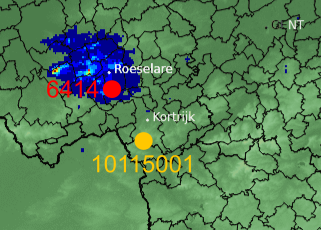}
        \caption{15h00}
        \label{fig:radar_image2}
    \end{subfigure}
    \begin{subfigure}[t]{0.48\textwidth}
        \centering
        \includegraphics[width=\linewidth]{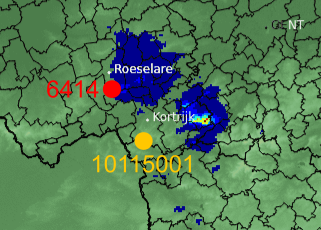}
        \caption{16h00}
        \label{fig:radar_image3}
    \end{subfigure}
    \hfill
    \begin{subfigure}[t]{0.48\textwidth}
        \centering
        \includegraphics[width=\linewidth]{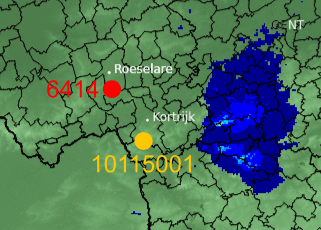}
        \caption{17h00}
        \label{fig:radar_image4}
    \end{subfigure}
    \caption{Precipitation radar imagery showing the eastward progression of a convective cell on June 23, 2023. The storm system passes directly over station \texttt{6414} between 15:00 and 16:00, while station \texttt{10115001} remains unaffected, illustrating the highly localized nature of convective precipitation events.}
    \label{fig:rain_radar}
\end{figure}

We filled in gaps in data from station \texttt{6414} during the rainfall window, but the completion disregarded the local cooling and produced estimates more consistent with neighboring stations that did not experience any precipitation event.

This situation shows that relying on temperature alone does not fully explain what happens when localized weather creates disturbances. If two weather stations are close together, they might be on either side of a small area with heavy rain. This is illustrated in \Cref{fig:rain_radar} which shows hourly precipitation accumulation observed by the weather radars on June 23, 2023 at 14:00, 15:00, 16:00 and 17:00. These radar images show that these precipitation events spread eastward, passing through station \texttt{6414} but not reaching station \texttt{10115001}.

Addressing these situations would require incorporating additional meteorological variables or mechanisms designed to detect and preserve such local anomalies, and to treat them separately during the imputation.

\section{Conclusions}
\label{sec:conclusions}

We have presented a novel approach called Graph-Regularized Riemannian Trust-Region Matrix Completion (GR-RTRMC). The method extends RTRMC by incorporating graph regularization to exploit structural relationships in the data. Experiments on meteorological and collaborative filtering datasets demonstrate the effectiveness of our approach.

In the meteorological case, highly localized phenomena such as scattered precipitation events complicate the imputation for matrix completion-based methods. These local disturbances violate the spatial coherence assumptions underlying low-rank models, highlighting the inherent limitations of approaches that rely solely on matrix completion techniques.

Several directions for future work emerge from this study. Incorporating additional meteorological variables (e.g., precipitation, humidity, solar radiation) into the model could help detect and handle localized weather events more effectively. Developing mechanisms to automatically detect anomalous periods and reduce the confidence assigned to observations during such events could improve robustness without manual intervention. Finally, alternative graph construction strategies could be explored to further improve completion accuracy.

\backmatter

\bmhead{Acknowledgements}
Computational resources have been provided by the supercomputing facilities of the Université catholique de Louvain (CISM/UCL) and the Consortium des Équipements de Calcul Intensif en Fédération Wallonie Bruxelles (CÉCI) funded by the Fond de la Recherche Scientifique de Belgique (F.R.S.-FNRS) under convention 2.5020.11 and by the Walloon Region.

\section*{Statements and Declarations}

\bmhead{Funding}
This work was funded by the Fonds de la Recherche Scientifique--FNRS under Grant no T.0001.23.

\bmhead{Competing Interests}
The authors have no competing interests to declare that are relevant to the content of this article.

\bmhead{Availability of data and material}
The code that produced the experimental results is available at \url{https://github.com/bloucheur/CAM-2026}.

\bibliography{references}

\end{document}